\documentclass[letterpaper, 10 pt, conference]{ieeeconf}
\IEEEoverridecommandlockouts

\usepackage{amssymb}
\usepackage{amsmath}
\usepackage[nolist]{acronym}
\usepackage{bm}
\usepackage{booktabs}
\usepackage{color}
\usepackage{cite}
\usepackage{graphicx}
\usepackage{multirow}
\usepackage{tabularx}
\usepackage{siunitx}
\usepackage[normalem]{ulem}
\usepackage{url}

\newacro{dl}[DL]{Deep Learning}
\newacro{fov}[FoV]{Field of View}
\newacro{lc}[LC]{Logistic Classifier}
\newacro{lstm}[LSTM]{Long Short-Term Memory}
\newacro{ml}[ML]{Machine Learning}
\newacro{mlp}[MLP]{Multilayer Perceptron}
\newacro{nn}[NN]{artificial Neural Network}
\newacro{hri}[HRI]{Human-Robot Interaction}
\newacro{rf}[RF]{Random Forest}
\newacro{rnn}[RNN]{Recurrent Neural Network}
\newacro{shri}[SHRI]{\emph{Socially-accepted} Human-Robot Interaction}
\newacro{ssl}[SSL]{Self-Supervised Learning}
\newacro{auroc}[AUROC]{Area Under Receiver Operating Curve}

\graphicspath{{./figures}}

\title{\LARGE \bf
Predicting the Intention to Interact with a Service Robot:\\the Role of Gaze Cues
}

\author{Simone Arreghini, Gabriele Abbate, Alessandro Giusti, and Antonio Paolillo
\thanks{This work was supported by the European Union through the project SERMAS, by the Swiss State Secretariat for Education, Research and Innovation (SERI) under contract number 22.00247, and by the Swiss National Science Foundation (grant n. 213074).}
\thanks{All the authors are with Dalle Molle Institute for Artificial Intelligence (IDSIA), USI-SUPSI, Lugano, Switzerland {\tt name.surname@idsia.ch}\linebreak
~
\linebreak
\vspace{-0.75mm}
\noindent\fbox{\footnotesize\begin{minipage}{0.99\textwidth}\copyright 2024 IEEE.  Personal use of this material is permitted. Permission from IEEE must be obtained for all other uses, in any current or future media, including reprinting/republishing this material for advertising or promotional purposes, creating new collective works, for resale or redistribution to servers or lists, or reuse of any copyrighted component of this work in other works.
\end{minipage}}
\vspace{-1.5cm}%
}%
}

\begin{document}

\maketitle
\thispagestyle{empty}
\pagestyle{empty}

\begin{abstract}
For a service robot, it is crucial to perceive as early as possible that an approaching person intends to interact: in this case, it can proactively enact friendly behaviors that lead to an improved user experience.  We solve this perception task with a sequence-to-sequence classifier of a potential user intention to interact, which can be trained in a self-supervised way. Our main contribution is a study of the benefit of features representing the person's gaze in this context.  Extensive experiments on a novel dataset show that the inclusion of gaze cues significantly improves the classifier performance (AUROC increases from \SI{84.5}{\percent} to \SI{91.2}{\percent}); the distance at which an accurate classification can be achieved improves from \SI{2.4}{\meter} to \SI{3.2}{\meter}. We also quantify the system's ability to adapt to new environments without external supervision. Qualitative experiments show practical applications with a waiter robot.
\end{abstract}
\section{Introduction}\label{sec:intro}

The increasing use of service robots demands safe and efficient interactions with humans, which challenges scientists to build machines with social skills.
\ac{hri} applications are demonstrating the potential of robots in providing relevant services, such as home assistance~\cite{Zachiotis:icrb:2018}, reception~\cite{Lee:cscw:2010}, and hospitality~\cite{Tuomi:chq:2021}. 
However, much research effort still needs to be carried out to endow robots with advanced perception capabilities for predicting the behavior of nearby people.
Consider, for instance, the scenario where a service robot has to assist the customers of a shop, or the guests of an hotel, who ask for information.
In these circumstances, it is desirable that robots can understand human intentions on their own, well before the interaction actually starts.
In this way, the assisting robot can enact friendly approaching behaviors so that even the hesitant, shy, or skeptical user can be well accommodated.
To make this possible, it is crucial to provide the robot with the ability to predict the intention to interact of potential nearby users.
In this very initial phase of the interaction, where the person is far away from the robot, possibly in a cluttered and noisy environment, nonverbal communication, such as body language or proxemics~\cite{Urakami:thri:2023},~\cite{takayama_influences_2009}, plays a crucial role.

\begin{figure}
    \centering
    \vspace{2mm}
    \includegraphics[width=0.86\columnwidth]{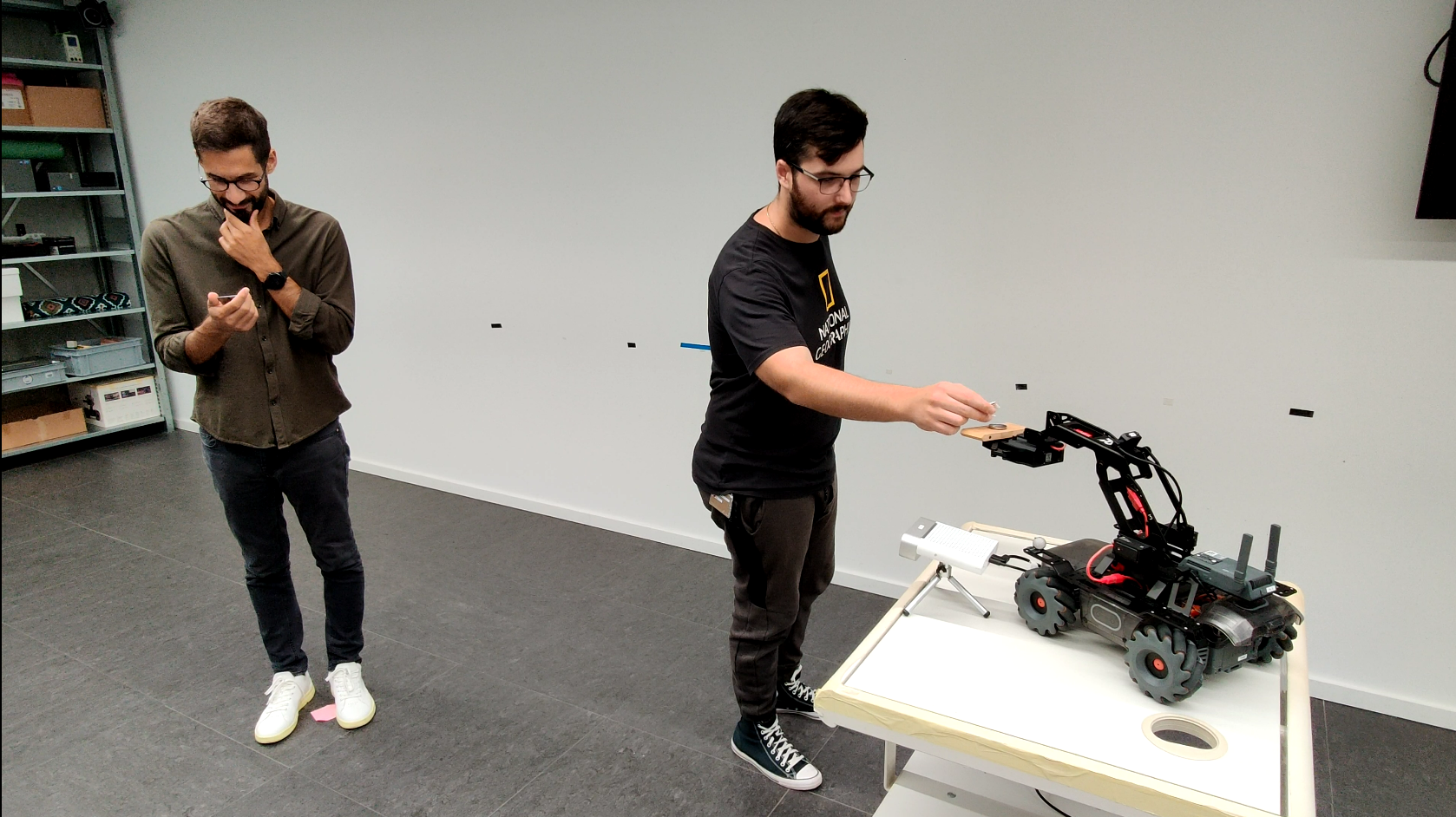}
    \caption{
A service robot predicts if a nearby person intends to interact, so to proactively enact a friendly behavior.
}
    \label{fig:setup}
\end{figure}
This work uses a self-supervised approach that allows a service robot to predict, for any person in its \ac{fov}, if that person intends to interact.  
The approach is based on a sequence-to-sequence classifier that, given a human's pose and gaze, continuously predicts whether it is going to eventually interact. 
It is self-supervised because, after the subject eventually moves away, the robot can, without external supervision: reconsider the entire sequence; label it with the corresponding ground truth, i.e. whether the subject actually interacted or not; add the labeled sequence to the classifier's training set; retrain its classifier.

In a previous work~\cite{Abbate:ras:2024}, we explored a self-supervised approach using features of the person's body motion. 
The \textbf{main contribution} of this work is to assess the improvement due to gaze cues.  
To this end, we develop a classifier that, given sensing data, predicts mutual gaze, i.e. when someone is actively looking at the robot camera.
We record in 3 distinct locations a novel dataset including $84$ positive and $105$ negative sequences with a service robot.
Experimental results show excellent prediction performance when a \ac{rnn} uses as input both the subject's pose and gaze. 
Most notably, gaze information enables the classifier to achieve high accuracy ($95.2\%$) at an average subject distance of $3.2$~m, compared to a much shorter distance ($2.4$~m) at which a classifier without gaze cues achieves the same accuracy.  
Additional experiments quantify the approach's ability to adapt to new environments in a self-supervised way and demonstrate human-friendly behaviors that a waiter robot can enact when using the proposed model.

The remainder of the paper is organized as follows: Sec.~\ref{sec:related_work} discusses the related literature, whereas our approach is described in Sec.~\ref{sec:model}.
Section~\ref{sec:setup} introduces the experimental setup used for the evaluation, whose results are presented in Sec.~\ref{sec:results}.
Finally, concluding remarks are reported in Sec.~\ref{sec:conclusions}.

\section{Related work}\label{sec:related_work}

Nonverbal communication is fundamental component in the context of \ac{hri}, for both humans and robots~\cite{Gasteiger:ijsr:2021,Saunderson:ijsr:2019}.
Nonetheless, the extraction of useful information about the users' behavior is not an easy task in \ac{hri}, especially when dealing with nonverbal communication~\cite{Rios:ijsr:2015}.
A body of work aims at estimating the human intention, e.g. in the context of navigation~\cite{Agand:icra:2022}, collaborative tasks~\cite{Belardinelli:iros:2022,Vinanzi:icdler:2019}, or for social behavior interpretation~\cite{Zaraki:icrm:2014,Gaschler:iros:2012, del2020you}. 
An important role, in accurately predicting the human intention to interact, is played by the information enclosed in the user gaze~\cite{belardinelli_gaze-based_2023,admoni_social_2017}.
The information about a person's gaze can be summarized into two types: gaze direction estimation and mutual gaze detection. 
Accurate enough gaze tracker methods can be found in the literature since a long time; such methods try to solve the problem of gaze vector regression. 
Some \ac{dl} based trackers~\cite{zhang2015appearance,krafka2016eye,zhang2017s} were developed in the past, however, most of the time they fall short in tracking at high distances, which is a problem the context of \ac{hri}. 
Long-range methods have been proposed~\cite{hennessey2012long,cho2013long} but they usually require cumbersome 
hardware that is difficult to mount on mobile robots. 
Recent developments~\cite{zhang2022gazeonce} showed promising performances in solving this problem, however, user implementation is not yet available, with training and deployment requiring complex setup. 
A much simpler task is the problem of detecting mutual gaze, which is generally defined as mutual eye contact between individuals, or between a person and a robot camera sensor. 
It has been studied with great results both outside~\cite{chong2020detection} and inside the robotics community~\cite{lombardi2022toward}. 
However, these methods again fall short in tracking at high distances.
Specifically to the \ac{hri} domain, many approaches base the intention recognition only on gaze cues~\cite{belardinelli_gaze-based_2023} or combined with other features like body motion cues~\cite{Brenner:roman:2021,Vaufreydaz:ras:2016,Kato:hri:2015,bi_method_2023, gaschler2012social}.
Our approach extends, adding mutual gaze cues, what has been proposed in~\cite{Abbate:ras:2024} where we devised a self-supervised algorithm for detecting the intent to interact of a human with a robot leveraging only body motion cues. 
It is worth mentioning that in the \ac{dl} literature, the term \ac{ssl} denotes the practice of using pretext tasks~\cite{Jing:pami:2020,Doersch:iccv:2017,Nava:ral:2022} for learning representations from big unlabeled data.
Instead, we refer to the meaning used in the robotics literature since the mid-2000s: self-generating supervision by leveraging data collected in previous experiments by the robot's own sensors, a paradigm successfully applied to several robotic applications, see e.g.~\cite{Nava:ral:2021,Lookingbill:jcv:2006,Hadsell:jfr:2009}.
\section{Model}\label{sec:model}

\subsection{Problem formulation and solution approach}\label{sec:problem}

Consider a service robot stationary in a public space, awaiting a possible assistance request from a user.
The robot is assumed to be equipped with exteroceptive sensing capabilities, e.g. provided by an RGB-D sensor.
The representative frame of the robot is chosen at its camera sensor frame and is denoted with ${\cal F}_s$.
The subject freely moves in the environment, i.e., they can randomly enter or exit the scene, and possibly interact with the robot.
The information about the person's behavior is described by properly chosen body frames and facial landmarks.
The body frames of interest are one located in the middle of the person's chest (denoted with ${\cal F}_c$) and on the person's head (${\cal F}_h$).
The 3D poses of the frame ${\cal F}_c$ and ${\cal F}_h$ w.r.t ${\cal F}_s$ are denoted with $\bm{p}_c \in \mathbb{R}^{9}$ and $\bm{p}_h\in \mathbb{R}^{9}$, respectively\footnote{Following best practices in machine learning, each Euler angle of the orientation representation is encoded as its $\sin$ and $\cos$ functions to avoid discontinuities around zero (in this case the pose size is $9$).}.
We assume that such metric information, indicative of the proxemics of the subject, is measurable by the robot sensor.
Furthermore, we also assume that the camera RGB images allow the detection of facial landmarks, which mainly consist of the projected locations, on the image plane, of specific points of interest on the person's face. 
Multiple facial landmarks are detected at once, each of which is denoted with the variable $\bm{\rho}_i\in\mathbb{R}^3$.
Each landmark contains the 2D point coordinates on the image with an additional component corresponding to the depth of the landmark w.r.t. the face center of gravity.
The facial landmarks, together with the body information of the subject's chest and head, are fed to a pre-trained \emph{mutual gaze classifier} that outputs a score $\hat{m}$ representing the probability that the subject is looking at the robot.
Finally, the subject's intention to interact is indicated with the boolean $y$.

\begin{figure}[t]
\vspace{2mm}
    \centering
    \includegraphics[width=0.94\columnwidth]{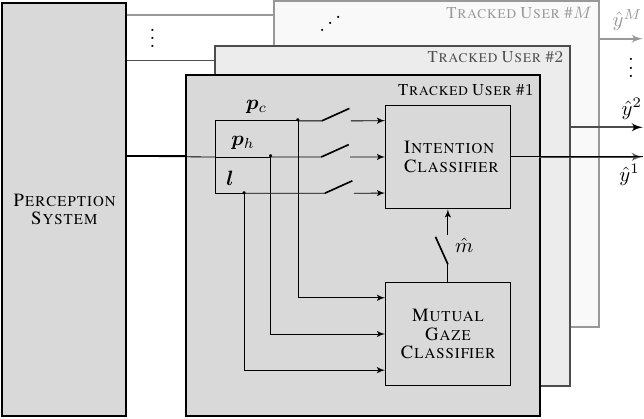}
    \caption{System architecture.}
    \label{fig:architecture}
\end{figure}
The perception problem tackled in this work is the prediction of a potential user's intention to interact with the robot, by monitoring their body motion, facial landmarks, the mutual gaze, or a combination of them. 
The block diagram of the proposed solution is shown in Fig.~\ref{fig:architecture}: our system is a classifier that, given the information about a potential user, provides an estimate $\hat{y}$ of its probability of interaction with the robot.
The approach relies on ($i$) existing modules providing information on people motion, and ($ii$) another specifically designed classifier, which computes the mutual gaze.
Multiple subjects are handled, as different instances of the approach can be instantiated in parallel.
In our analysis, we compare different combinations of input taken from the robot's lower-level perception pipeline, as well as different implementations of the classifier.
In this section, we present the data structure and the feature sets required to train our intention to interact prediction model.

\subsection{Dataset}\label{sec:dataset_def}
The training dataset ${\cal D}$ is organized into \emph{sequences}, each one related to a subject who appears in the \ac{fov} of the sensor.
Each whole sequence is given a boolean label, which reflects the occurrence of an interaction (true) or not (false).
The sequences are composed of \emph{frames}, i.e. sample data at each timestamp; frames belonging to the same sequence share the same label.
More in detail, the dataset has this shape:
\begin{equation}
    {\cal D} = \left\{\bm{f}_{i,j}, y_j \right\}_{i=1,j=1}^{N_j,S}
    \label{eq:dataset}
\end{equation}
where $S$ is the number of sequences, and $N_j$ is the number of frames in $j$-th sequence.
The features vector $\bm{f}_{i,j}$ contains the information about the person at the $i$-th frame of the $j$-th sequence; our study compares different choices of feature sets, as presented in Sec.~\ref{sec:features}.
The variable $y_j$ denotes the label for the $j$-th sequence.  
Note that, with the assumption that the robot is able to sense when an interaction occurs, any given past sequence can be labeled automatically using the robot's own experience, without any external supervision.

\subsection{Feature sets}\label{sec:features}

In our study, we analyze how the choice of features (introduced in Sec.~\ref{sec:problem}) impacts the prediction of the intention to interact with the robot.
The first and most simple set considers only the subject's chest pose:
\begin{equation}
    \bm{f}_\text{C} = \bm{p}_c \in \mathbb{R}^{9}
\end{equation}
The second set contains also the pose of the subject's head and actually serves as a baseline in our comparison since it is very similar to what has been used in~\cite{Abbate:ras:2024}:
\begin{equation}
    \bm{f}_\text{CH} = \left(\bm{p}_c^\top, \bm{p}_h^\top \right)^\top \in \mathbb{R}^{18}
\end{equation}
In the third set, we only consider the estimate of the mutual gaze information, as provided by the corresponding classifier: 
\begin{equation}
    \bm{f}_\text{M} = \hat{m} \in [0,1].
\end{equation}
The fourth one combines spatial cues with mutual gaze:
\begin{equation}
    \bm{f}_\text{CHM} = \left(\bm{p}_c^\top, \bm{p}_h^\top, \hat{m}  \right)^\top \in \mathbb{R}^{19}.
\end{equation}
The last one considers also the facial landmarks:
\begin{equation}
    \bm{f}_\text{FULL} = \left(\bm{p}_c^\top, \bm{p}_h^\top, \hat{m}, \bm{l}^\top \right)^\top \in \mathbb{R}^{19+3n}
\end{equation}
where $\bm{l} = (\bm{\rho}_1^\top, \dots, \bm{\rho}_n^\top)^\top \in \mathbb{R}^{3n}$ is the vector containing $n$ detected facial landmarks.

\section{Experimental setup}\label{sec:setup}
\subsection{Robot, sensing and perception}
In our work, we use a small wheeled omnidirectional robot (DJI RoboMaster EP\footnote{{https://www.dji.com/ch/robomaster-ep-core}}).
Furthermore, we use the Azure Kinect RGB-D sensor\footnote{https://learn.microsoft.com/en-us/azure/kinect-dk/system-requirement}, streaming images at a resolution of $4096\times3072$~pixels, with a nominal horizontal and vertical \ac{fov} equal to $90^\circ$ and $59^\circ$, respectively.
In our setup the camera sensor is placed in the proximity of the robot, at a height from the ground of around $1$~m (as in Fig.~\ref{fig:setup}), which is the typical height of robots designed for \ac{hri}, see e.g. Tiago by PAL Robotics~\cite{Pages:iros_ws:2016} or Pepper by Aldebaran~\cite{Pandey:ram:2018}. 
The sensor's SDK enables body frame tracking of multiple subjects simultaneously, directly providing us with the measurement of $\bm{p}_c$ and $\bm{p}_h$.
Facial landmarks for each subject are extracted with the  MediaPipe\footnote{https://developers.google.com/mediapipe/solutions} library.  
Its \emph{face mesh} module returns the 2D location on the image plane of salient face points, and the corresponding predicted depth relative to the center of mass of the face.
The facial landmarks originally expressed using image-normalized coordinates, are first centered around the landmarks' center of gravity and then normalized such that the mean landmark vector has unitary size. 
Mediapipe provides $478$ landmarks by default, from which we select $n=39$ points representative of major structures (mouth, eye corners, irises, nose, face contours), whose normalized coordinates are concatenated in vector $\bm{l}$.
\subsection{Dataset collection}\label{sec:dataset}
The dataset was collected by deploying the system and recording multiple sequences.  For each sequence, a person enters the 
robot \ac{fov} and either interacts with it before leaving (positive sequence) or simply transits near the robot without interacting (negative sequence). 
An interaction consists of taking a chocolate treat that RoboMaster holds.
The entire dataset contains a total of $4946$ frames and $189$ sequences ($84$ positive, $189$ negative).  
Data was split between $3$ different environments, using different subjects.

\emph{Lab}: $92$ sequences ($33$ positive, $59$ negative) recorded in a mostly empty laboratory, size $9\times9$~m, with a single subject visible at a time.  Only for this sequence, the ground truth of the subject's gaze is also collected and used to test the mutual gaze classifier.
\emph{Office}: $42$ sequences ($15$ positive, $27$ negative) collected in a corridor of an office building, including multiple subjects simultaneously present in many frames.
\emph{Kids}: $55$ sequences ($36$ positive, $19$ negative), collected in a break area of an office building. Subjects are $10$ to $12$ years old kids; each kid was instructed to either traverse the break area (without providing specific instructions regarding the robot) or stop and grab the chocolate treat from the robot.

All the participants or their tutors (for the kids), signed a consent form; all data is kept private; the experiments were approved by the local institution's ethical committee.
\subsection{Classifier architecture}\label{sec:architecture}

We expect that the motion dynamics of body frames, facial landmarks, and the temporal evolution of the mutual gaze are important cues to predict whether a given person intends to interact with the robot.  
To capture these dynamics, we use a recurrent \ac{lstm}~\cite{LSTM} neural network as a stateful sequence-to-sequence classifier.
We use the implementation available in the PyTorch\footnote{https://pytorch.org/} library. 
All the LSTM models have $2$ layers, whereas the hidden state dimension varies to accommodate the different sizes of the feature sets (from $10$ for $\bm{f}_\text{M}$ to $65$ for $\bm{f}_\text{FULL}$).
To offer a more complete analysis, we compare the \ac{lstm} performance against a simpler, stateless model, i.e. a \ac{rf} classifier implemented using the scikit-learn package.

All the models are evaluated using a 5-fold stratified cross-validation strategy; folds are computed by splitting the set of sequences in training and testing sets, such that all frames for a given sequence stay in the same set. 
\subsection{Mutual gaze classifier}

In our setup, some feature sets include mutual gaze information through the variable $\hat{m}$ (see Sec.~\ref{sec:features}).
Off-the-shelf algorithms for mutual gaze estimation~\cite{lombardi2022toward} are not designed for distances larger than $1$~m; therefore, we trained an ad-hoc mutual gaze classifier tailored to our sensing setup.
To this end, we collected a dataset $\mathcal{D}_\text{gaze}$ in which subjects stood in front of the robot in predefined positions, arranged in such a way to cover the sensor's \ac{fov}.  For each position, the subject moves their head and torso while alternating periods of looking at the robot and periods of looking elsewhere; ground truth is collected by the subject himself, by keeping a button pressed when and only when the gaze is on the robot.
More in detail, with reference to Fig.~\ref{fig:architecture}, our mutual gaze module takes as input $\bm{p}_c$, $\bm{p}_h$ and $\bm{l}$. 
Data thus collected produced a training set of $12849$ samples.
Since we expect that face landmark dynamics are not useful to predict mutual gaze, we rely on a simple stateless \ac{rf} classifier, which can provide good robustness while being very lightweight.
More details about our mutual gaze classifier can be found in~\cite{Arreghini:hri:2024}.
\begin{figure}[t]
    \centering
    \includegraphics[width=\columnwidth]{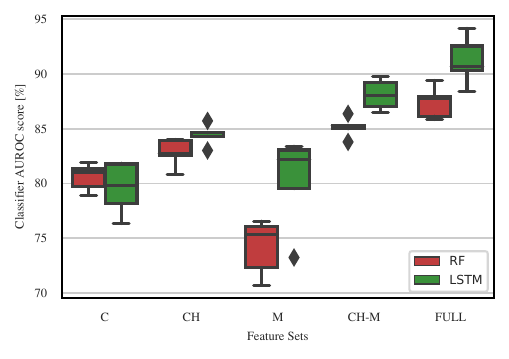}
    \caption{AUROC of \ac{rf} and \ac{lstm} classifiers with the different feature sets.}
    \label{fig:rf_vs_lstm}
\end{figure}
\section{Experimental results}\label{sec:results}

\subsection{Mutual gaze estimation performance}

\begin{figure}[t]
    \centering
    \includegraphics[width=\columnwidth]{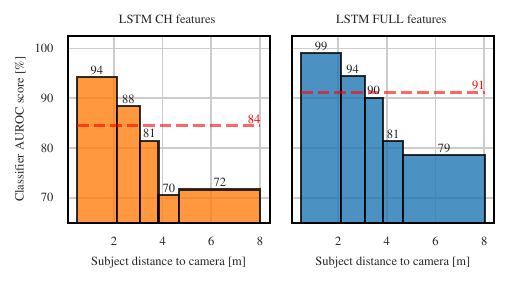}
    \caption{AUROC for the LSTM using $\bm{f}_\text{CH}$ (left) and $\bm{f}_\text{FULL}$ (right) for different human-robot distance quantiles.
    }
    \label{fig:distance_quantiles}
\end{figure}
When evaluated using 5-fold cross-validation on $\mathcal{D}_\text{gaze}$, the mutual gaze classifier yields an \ac{auroc} value of $91.9\%$, an accuracy of $83.3\%$, and an average precision of $86.3\%$.
We further verify the performance of the classifier in our relevant environments by evaluating it on the frames of the interaction dataset $\mathcal{D}$ (Sec.~\ref{sec:dataset}) that were collected in the \emph{Lab} environment, for which we have ground truth for subject gaze.
On this testing set, the mutual gaze classifier yields: $90.4\%$ AUROC, $82.4\%$ accuracy, $82.2\%$ average precision.
\begin{figure*}[t]
    \centering
    \includegraphics[width=\textwidth]{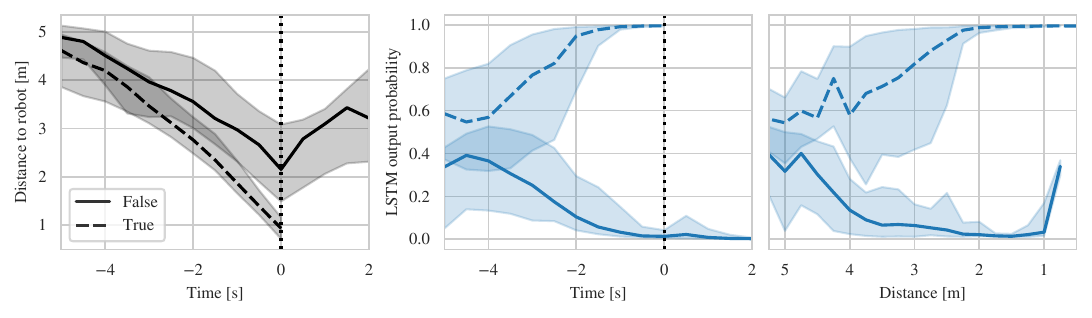}
    \caption{Median distance to the robot (left) and median predicted probability of interaction (center) as a function of time. Time $t=0$ is defined for each sequence as the moment when the subject either interacts, for positive sequences (dashed line), or the moment in which the subject is closest to the robot, for negative sequences (continuous line). The rightmost plot reports the predicted probability of interaction as a function of distance to the camera, ignoring negative samples with $t>0$. Shaded areas represent the interquartile range.}
    \label{fig:sequence_plot}
\end{figure*} 
\begin{figure*}[!ht]
    \centering
    \includegraphics[width=\textwidth]{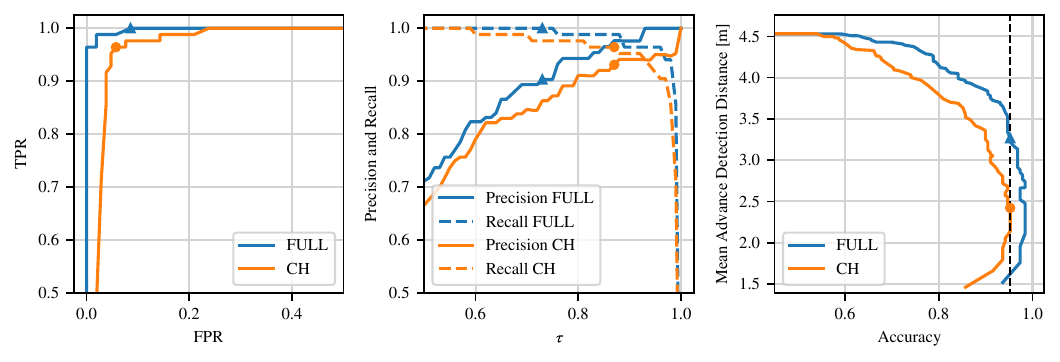}
    \caption{Sequence-level performance metrics for the LSTM approach with (blue) and without (orange) gaze and face landmark features. Left: ROC curve. Center: Precision and Recall as a function of threshold $\tau$. Right: Mean Advance Detection Distance (vertical axis) vs. achieved accuracy (horizontal axis) for different values of $\tau$. The orange circles denote a threshold value of $\tau=0.87$ for the baseline model set to achieve maximum accuracy. Conversely, the blue triangles denote a threshold value of $\tau=0.73$ needed by our model to display the same level of accuracy.}
    \label{fig:sequence_wise_metrics}
\end{figure*}

\subsection{Intention to interact: frame-level performance}
We first evaluate the algorithm performances frame-by-frame using the AUROC metric, comparing the two different model architectures (\ac{rf} and \ac{lstm}) introduced in Sec.~\ref{sec:architecture}.
The plot in Fig.~\ref{fig:rf_vs_lstm} shows that the stateful \ac{lstm} classifiers consistently outperform the simpler stateless \ac{rf} counterparts for every input feature set.
In the following, we focus our comparison on two models: ($i$) the LSTM model using $\bm{f}_\text{CH}$, which we refer to as the \emph{baseline}, as it is most representative of the model introduced in~\cite{Abbate:ras:2024}; and ($ii$) the LSTM model using $\bm{f}_\text{FULL}$ that we refer to as \emph{our model}.  The latter differs from the former since it includes gaze and face landmark cues.
The plot in Fig.~\ref{fig:rf_vs_lstm} shows that adding gaze information significantly improves classifier performance, increasing AUROC from $84.5\%$ for the baseline to $91.2\%$ for our model. 

Fig.~\ref{fig:distance_quantiles} reports an  experiment in which we split the testing data into $5$ distance quintiles, and evaluate the classifier separately on each.  All reported AUROC values are significantly greater than $0.5$, which indicates that, even among subjects that are at approximately the same distance from the robot, the classifier is effective at differentiating those who are likely to interact and those who are not; i.e., even though subject distance from the robot is a powerful feature, it is not the only aspect that is considered by the models.
The figure further shows that the improvement of the performance introduced by the contribution of the gaze is uniform across the whole range of subject-robot distances. 
In the bin of the closest distance (the one from $0.48$ to $2.10$~m), the gain in performance due to the additional gaze information and facial landmarks is about $5\%$. 
This improvement increases to $9\%$ for the intermediate bin (containing distances from $3.08$ to $3.83$~m), and to $7\%$ for the farthest distances (above $4.68$~m).

Fig.~\ref{fig:sequence_plot} shows the statistics from positive and negative sequences for the whole dataset. 
In the left and center plots, all sequences are temporally aligned in such a way that $t=0$ denotes the time of interaction, in the case of positive sequences, and the time in which the subject is at the closest distance from the robot, in case of negative sequences. 
We observe from the left plot that negative sequences reach, on average, a distance from the robot of $1.6$~m before moving further. 
The center plot shows that at $t=0$ (vertical dotted line), the model yields very sharp predictions.
The distribution of the output probabilities for positive and negative sequences start diverging at $t=-4$~s, and clearly separate at $t=-3$~s.
The rightmost plot reports the same data but with distance to the robot on the horizontal axis. 
Negative sequences that reach distances below $1$~m are few, so the rightmost data is noisy.

\subsection{Intention to interact: sequence-level performance}

We now report the performance of our models when evaluating them at the level of entire sequences.  For each sequence, we simulate that the model is applied to each frame, and when exceeding a threshold $\tau$, the robot takes an irreversible decision to enact a given behavior (e.g. facing, approaching or greeting the user).  Negative sequences in which the output probability never exceeds $\tau$ are true negatives (i.e., the robot correctly ignored a non-interacting subject); positive sequences in which the output probability exceeds $\tau$ for at least a single frame are true positives; only for true positives, from the earliest frame whose classifier output exceeds $\tau$, we compute the \emph{advance detection time} (i.e. the amount of anticipation) and \emph{advance detection distance} (i.e. the distance of the user when the robot reacted). False negatives denote sequences for which the robot did not react to an interacting user; false positives denote sequences in which the robot incorrectly reacted to a non-interacting subject. Given these definitions, we compute sequence-level metrics: False Positive Rate (FPR), True Positive Rate (TPR), precision, recall and accuracy.
Figure~\ref{fig:sequence_wise_metrics} reports these metrics for both the baseline model (which uses $\bm{f}_\text{CH}$), in orange, and our model (that relies on the more complete information contained in $\bm{f}_\text{FULL}$), in blue.  
We observe that the latter outperforms the former in all metrics, regardless of the threshold value $\tau$. 
In particular, the center plot highlights that high threshold values are key to obtaining very high precision and recall performance, with our model consistently outperforming the baseline in both metrics. 
Nevertheless, the plot does not show the complete picture: high threshold values yield high performance because, in this case, the model does not commit to a decision until very late in the sequence, when most positive sequences yield very high probabilities; this behavior is not useful in practice. 
The right plot in Fig.~\ref{fig:sequence_wise_metrics} studies the trade-off between sequence classification accuracy, on the horizontal axis, and mean advance detection distance, on the vertical axis, controlled by $\tau$.
Low values of $\tau$ yield early, distant but inaccurate detections (top left). 
Increasing $\tau$ decreases the mean Advance Detection Distance but improves accuracy up to a maximum value; further increases of $\tau$ lead to a marked increase in false negatives, and negatively impact both Advance Detection Distance and Accuracy, as can be also seen from the drops in the recall value. 
The orange dots denote a threshold ($\tau = 0.87$) yielding maximum accuracy ($95.2\%$) for the baseline classifier, and the blue triangles denote the threshold ($\tau = 0.73$) needed to get to the same accuracy for our model.   At this threshold, our model yields a significantly better advance detection distance ($3.27$~m) w.r.t. the baseline ($2.42$~m): an improvement of $0.85$~m.

\begin{figure}[t]
    \centering
    \includegraphics[width=\columnwidth]{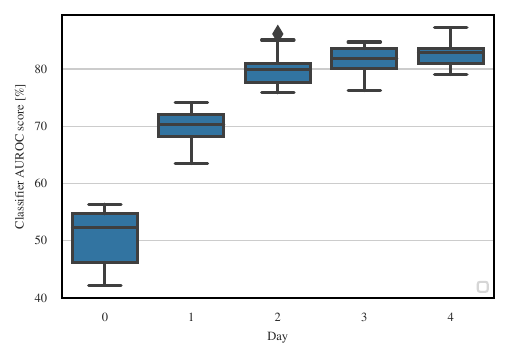}
    \caption{AUROC during self-supervised adaptation to a new environment.}
    \label{fig:ssl}
\end{figure}

\subsection{Self-supervised adaptation to new environments}

To test the self-supervised ability of the proposed approach, we consider the situation of a robot with a model that is pre-trained in two environments (\emph{Lab} and \emph{Office}) and then deployed in a new one (\emph{Kids}), which has a different layout and where subjects behave differently.
We use $1/5$ of the sequences in the deployment environment as our fixed testing set. 
The remaining sequences of the deployment environment are split into four groups (\emph{Day 1} to \emph{Day 4}) which are assumed to be collected by the robot in a self-supervised way during its first few days of deployment. 
Figure~\ref{fig:ssl} reports the frame-level AUROC computed on the fixed testing set:
day~0 represents the model trained only on the training data, which has not yet adapted to the new environment; the subsequent entries represent the classifier trained on data collected from the training environments plus the deployment environment up to day $n \in \{1,2,3,4\}$.
Results are reported for $10$ replicas, obtained by different random splits of the data in the deployment environment.
We observe that performance at day 0 is very poor (AUROC $\approx 0.5$), because the behavior of kids and the room layout is very different from the training data; performance increases as the robot accumulates additional experience.
\begin{figure}[!t]
    \vspace{2mm}
    \centering
    \includegraphics[trim={4cm 2cm 6cm 5cm},clip,width=0.47\columnwidth]{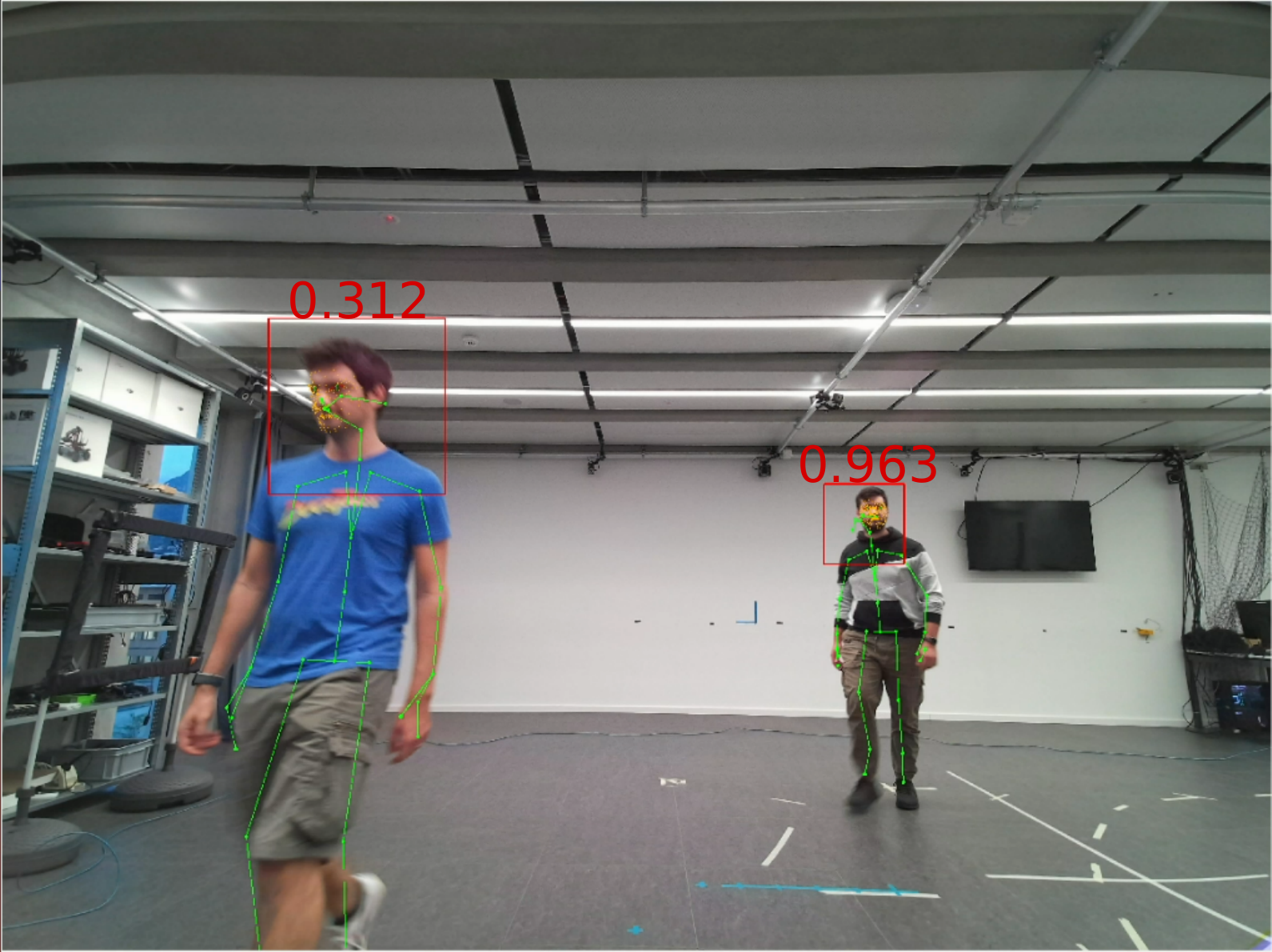}
    \hfill 
    \includegraphics[trim={9cm 2cm 1cm 5cm},clip,width=0.47\columnwidth]{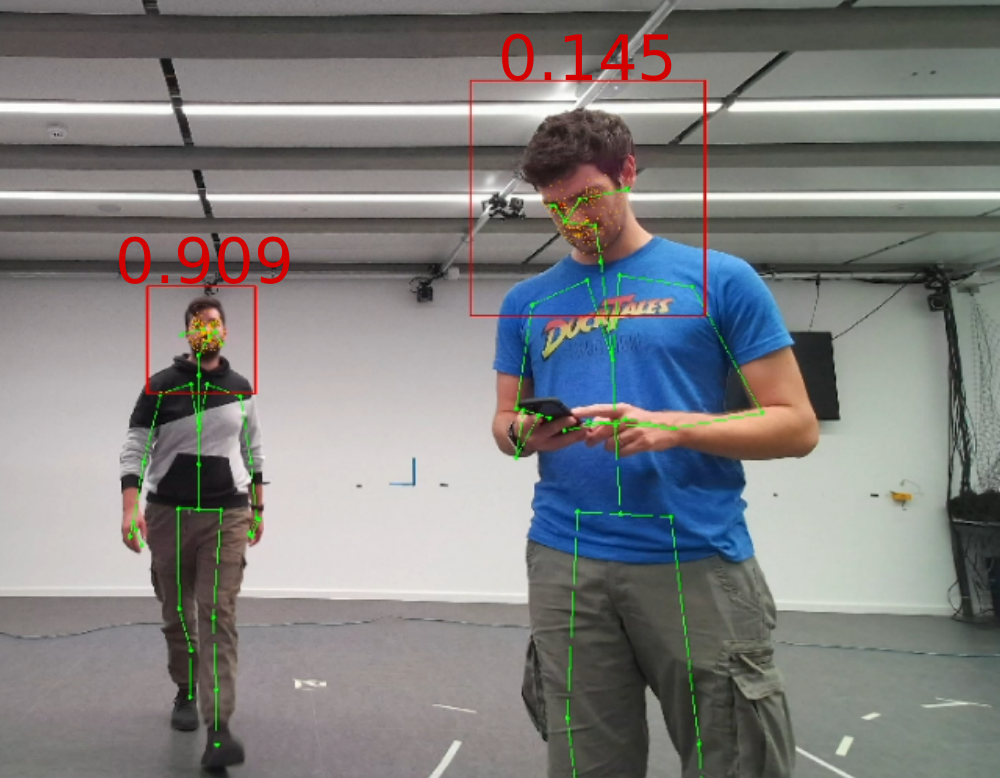}
    \caption{Two snapshots of the robotic experiments used to evaluate our framework in real-life scenarios.}
    \label{fig:real_world_test}
\end{figure}
\subsection{Evaluation of the intention to interact in a robotic task}

Our approach has been evaluated in real experiments. 
Two indicative snapshots of this experiment are shown in Fig.~\ref{fig:real_world_test}.
In the right image, a person is standing very close to the robot (the one with a predicted probability of $0.145$) but is not paying attention as he is looking at his phone.
Our predictor recognizes this situation, by providing a low predicted probability of interaction; the other subject in the same snapshot, instead, is walking and looking toward the robot and his predicted probability is high ($0.909$). 
Similar conclusions can be drawn by looking at the left snapshot. 
Here, instead of standing, the person not interested in interacting (the man in blue) is walking past the robot and is given a low probability ($0.312$), whereas the other person shows interest in interacting (the man in the background), with a predicted interaction probability of $0.963$.  

This experiment and other results can be seen in the video accompanying this paper.

\section{Conclusions}\label{sec:conclusions}
We have shown that gaze cues improve the prediction of a potential user's intention to interact with a service robot; in line with the literature, the presented system outperforms our previous work that only uses body motion cues.
The user's intention to interact prediction is an important perception task to enact proactive behaviors that yield effective and satisfactory interaction experiences for users.
The implementation consists of a classifier that takes as input the features of a person's body and gaze and outputs its probability of interaction.
We have compared two different architectures of classifiers (random forest and long-short term memory) with different feature sets.
Additional material is available at \url{https://github.com/idsia-robotics/intention-to-interact-detector-gaze}.

Future work will be devoted to testing 
our perception module in more challenging real-world social scenarios, such as the hall of a hotel or the entrance of a shopping mall, 
and experimenting with robot reaction behaviors. 
A user study will be carried out to evaluate the users' satisfaction level with service robots in real social contexts.

\addtolength{\textheight}{-2.7cm}  

\bibliographystyle{IEEEtran}
\bibliography{bibliography.bib}

\end{document}